\documentclass[journal]{IEEEtran}
\usepackage{amsmath, amsthm, amssymb, amsfonts} % Core math packages (amsmath must be before amsthm)
\usepackage{times} % Times font
\usepackage{graphicx} % Graphics inclusion
\usepackage{xcolor} % Color support
\usepackage{url} % Formatting URLs
\usepackage{cite} % Handling citations

\usepackage{stfloats} % Better control over floats (like \begin{figure*}[!b])
\usepackage{placeins} % Provides \FloatBarrier
\usepackage{array} % Extended array and column formatting
\usepackage{booktabs} % Professional-looking tables
\usepackage{multirow} % Multi-row cells in tables
\usepackage{diagbox} % Diagonal lines in table cells
\usepackage[caption=false,font=normalsize,labelfont=sf,textfont=sf]{subfig} % Subfigures/subtables (Modern replacement for subfigure, but need to check compatibility when using both)

\usepackage{algorithmic} % Algorithm formatting
\usepackage{algorithm} % Algorithm environment
\usepackage{verbatim} % Verbatim text environment

\usepackage{textcomp} % Additional text symbols
\usepackage{soul} % Highlighting text (for \hl)
\definecolor{darkgreen}{HTML}{006400} % Define dark green color
\sethlcolor{green} % Set highlight color to green

\allowdisplaybreaks % Allows math environments (like align) to break across pages

\usepackage[letterpaper, top=54pt, bottom=73pt, left=45pt, right=45pt, nohead, nofoot]{geometry}

\title{Empirical Pedestrian Safety Assessment in a Mobile Robot Using a Predictive Social Force Model}

\author{Alireza Jafari, Yun-Hao Tsai, and Yen-Chen Liu~\IEEEmembership{Senior Member,~IEEE}
\thanks{This work was supported in part by the National Science and Technology Council (NSTC), Taiwan, under Grants NSTC 114-2628-E-006-010 and NSTC 114-2218-E-006-021; and in part by the Higher Education Sprout Project, Ministry of Education, to the Headquarters of University Advancement at National Cheng Kung University (NCKU).}
\thanks{National Cheng Kung University ethics committee reviewed and approved all the study procedures (IRB Approval No. \textcolor{black}{NCKU~HREC-E-114-0870-2).}}
\thanks{The authors are with the Department of Mechanical Engineering, National Cheng Kung University, Tainan 70101, Taiwan (e-mail:~alireza.jafari.110@gmail.com; sean901109@gmail.com; yliu@mail.ncku.edu.tw).}}

\IEEEpubid{}
\begin{document}
\maketitle
\begin{abstract}
Mobile robots are going to share the sidewalks with pedestrians.
They must ensure their objective safety and respect the walkers' subjective safety/comfort.
Computationally efficient Social Force Models (SFM) present interpretable solutions for real-time robot navigation in dynamic crowds. 
Recent explorations of Projected Time-to-collision (PTTC) integration into SFM variants, for example, PTTC-based SFM (TSFM), improve safety metrics.
But the effect of predictive variants is unclear.
We introduce Predictive SFM (PSFM) and Predictive TSFM (PTSFM) by integrating predicted social force vectors over a finite time horizon.
The paper implements SFM, TSFM, PSFM, and PTSFM on a nonholonomic mobile robot and performs experimental trials with volunteers attending a facing scenario.
We systematically study objective and subjective safety across the variants.
Minimum PTTC, average speed, minimum distance, lateral distance, and the maximum trajectory curvature benchmark the objective safety.
Likert scale post-interaction surveys assess subjective safety by marking comfort, smoothness, distance appropriateness, and speed suitability.  
We confirm that PTTC integration improves safety metrics.
The prediction contribution is limited and occasionally visible in some of the sub-metrics.
Some participants perceive smoother movements and safer speed behavior with predictive methods, but Mann-Whitney tests reveal no significant differences in subjective ratings. 
Therefore, PTTC-based navigation enhances safety, whereas the formulated prediction offers limited additional benefits in single-pedestrian scenarios.
\end{abstract}
\section{Introduction}\label{sec:int:main}
\IEEEPARstart{P}{edestrians} are going to share their public spaces with mobile robots.
Examples of the mobile robots entering public spaces are delivery robots on sidewalks, service robots in hospitals~\cite{Jung2020}, robotic bodyguards in shopping malls~\cite{Shehata2025}, and companion robots in leisure spaces~\cite{Jung2023}.
The integration of robots into public spaces challenges people's safety.
Two aspects of human safety are objective safety and subjective safety.
Objective safety quantifies the physical contact risks using measurable variables.
Subjective safety captures the human feeling of being safe and is evaluated through surveys.
An objectively safe robot may still be subjectively unsafe and threatening if its movements appear erratic or violate social norms.
Nevertheless, most research, such as ~\cite{Liu2023}, assumes pedestrians as dynamic objects and ignores the psychological component of safety.
\par
Social Force Models (SFM) are the primary model for predicting pedestrian movements in a crowd~\cite{Helbing2000}.
Extensions to micro-mobility vehicles like e-scooters~\cite{LIN2024 ,Liu2022-TITS, JAFARI2024-SIMPAT} and Segways~\cite{Dias2018} are other emerging research directions.
SFM is also a promising choice for real-time robot navigation~\cite{Shiomi2014, Truong2017} because of its advantages, such as low computational cost, interpretability, and versatility in handling unknown dynamic crowds without requiring extensive learning.
For example, Ægidius et al. augment SFM and implement it on a quadruped robot as a target-following human companion~\cite{AEgidius2024}.
Repiso et al. propose adaptive SFM navigators to guide a robot through a crowd as a member of a human subgroup~\cite{Repiso2020, Repiso2024}.
Recent SFM variants incorporate subjective safety metrics, for example, Projected Time-to-collision (PTTC), into the SFM navigators~\cite{JAFARI2024-3-SORO}.
The integration presents an opportunity to design algorithms that improve objective and subjective safety simultaneously~\cite{JAFARI2024-2-NATCOM}.
\par
Distance-based algorithms are popular in mobile robot pedestrian interactions.
For example, Wang et al. use a fixed desired relative distance to the walker in a human identification and following setup~\cite{Wang2025}.
Following the same pattern, mainstream research about human safety/comfort around mobile robots considers violation of personal space as the primary safety quantifier~\cite{Francis2025}; the closer you get to people, the more uncomfortable they will be.
They rely on Hall's classification of human personal space when interacting with other humans~\cite{Hall1963}.
\par
A problem with this approach is that the original work classifies a static space for human-human socialization, not for human-robot interactions.
Hence, Neggers et al. study the shape of the personal space during interaction with robots and provide discomfort contours~\cite{Neggers2022}.
Moreover, Hoang et al. consider dynamic social zones in a socially aware navigation framework~\cite{Hoang2023}.
Another issue with the approach is that it assumes a distance-based index and neglects other contributors, such as relative speed, TTC, or curvature.
While for pedestrians with similar and quasi-static walking speeds (1--1.5 m/s) the assumption may be acceptable, for mobile robots and other intelligent agents with wider speed ranges it is not enough~\cite{JAFARI2024-2-NATCOM}.
In addition, Greenberg et al. examine robot trajectory curvature influence on human subjective safety and report that people trust mild curvatures more than sharp or smooth trajectories~\cite{Greenberg2025}.
Time-to-collision (TTC) is another contributor to pedestrian safety~\cite{Jafari2023-2-IFAC}.
\par
TTC is a collision risk and near-miss metric for structured and primarily 1-dimensional road traffic.
Jafari and Liu propose PTTC for e-scooters on the 2-dimensional space of unstructured sidewalks~\cite{JAFARI2024-2-NATCOM} and for intelligent agents in public heterogeneous spaces~\cite{JAFARI2024-3-SORO}.
TTC integration into robot movement algorithms provides a sense of collision immediacy for the robot~\cite{Shahriari2022}.
Jafari and Liu show that PTTC integration into SFM and on nonholonomic mobile robots improves pedestrian objective and subjective safety~\cite{Jafari2026-TRO}.
\par
Predictive navigators forecast pedestrian trajectories and proactively adapt to dynamic crowds.
Chen et al. present an anticipative framework that estimates pedestrians' intentions~\cite{Chen2022-2}. 
Then, the robot optimizes its trajectory based on the predicted routes.
Pedestrian trajectory forecast and proactive adaptations smooth the navigation and ensure pedestrian safety/comfort.
But, does prediction integration in SFM variants improve objective and subjective safety?
While prediction may reduce collision risk, the corresponding human perception is unclear, and anticipatory behavior could appear more intelligent or overly cautious. 
\par
This paper proposes a prediction integration into SFM variants and systematically addresses its effects on objective and subjective safety.
We combine pedestrian and robot predicted positions with SFM and PTTC-based SFM (TSFM), assuming constant velocities for a finite time horizon, and formulate Predictive SFM (PSFM) and Predictive TSFM (PTSFM).
Through a series of trials, we evaluate the objective safety across the four algorithms using measurable objective safety metrics.
In addition, we statistically analyze the pedestrian subjective safety using survey responses to four key factors: interaction comfort, smoothness, distance appropriateness, and speed suitability.
The key finding is that PTTC integration significantly improves the safety. 
However, the prediction effect is marginal in some factors and statistically insignificant in others.
\par
Overall, the novelties of the paper are the integration of prediction into SFM and TSFM and the systematic and experimental evaluation of the objective and subjective safety. 
The objective and subjective safety evaluations in SFM and TSFM are a replication of our previous work~\cite{Jafari2026-TRO}.
The development, formulation, and safety studies and discussions regarding their predictive counterparts, PSFM and PTSFM, are new.
\par
The remainder is as follows.
Section~\ref{sec:back:main} provides SFM and PTTC background. 
Section~\ref{sec:theory:main} describes the four navigation variants. 
Section~\ref{sec:exp:main} details experimental setup.
Section~\ref{sec:res:main} presents results. 
Section~\ref{sec:con:main} concludes with future directions. 
\section{Background}\label{sec:back:main}
In this section, we briefly review SFM basics and the PTTC concept~\cite{JAFARI2024-3-SORO,JAFARI2024-2-NATCOM}.
In addition, the section sets the foundations for subsequent navigation algorithm design.
\subsection{Social force models}
\begin{figure}
\centering
\includegraphics[width=0.45\textwidth]{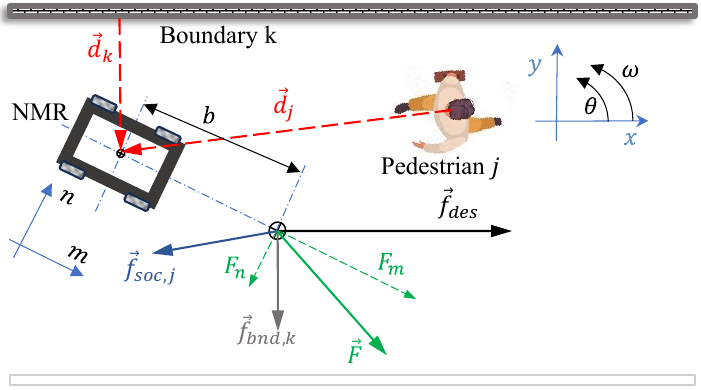}
\caption{The interaction geometry, the total force $\vec{F}$, and the projections $F_m$ and $F_n$ in local $m$-$n$ frame.
$\vec{F}$ is the linear superposition of boundary forces, social forces, and the desired force.
The robot moves forward in response to $F_m$ and rotates in response to $F_n$. 
The point of force application is a design choice. 
}
\label{fig:SFM_illustration}
\end{figure}
Originally, Helbing et al. designed SFM to simulate micro movements in a human crowd for macro pedestrian behavior predictions~\cite{Helbing1995, Helbing2000}.
Since then, SFM has been the primary model in pedestrian crowd behavior analysis~\cite{Moussaid2009}.
In addition, its extensions to human-controlled micro-mobility vehicles, like Segways~\cite{Dias2017,Dias2018} and e-scooters~\cite{Liu2022-TITS, JAFARI2024-SIMPAT}, provided insights for mixed crowd behavior on shared spaces. Mobile robots used SFM and its variations as a navigation algorithm, too~\cite{Shiomi2014, Kamezaki2022, Mac2025, Truong2017, Jafari2026-TRO}.
\par
SFM formalizes the pedestrians' movements, assuming they are the result of people's reactions to the superposition of certain psychological forces~\cite{Helbing1995},
\begin{equation}
\vec{F}=\vec{f}_{des}+\sum_{j=1}^{N}\vec{f}_{soc,j}+\sum_{k=1}^{B}\vec{f}_{bnd,k}\;,
\label{eq:SFMForce}
\end{equation}
where $\vec{F}$ is the total force exerted on the pedestrian.
$\vec{f}_{des}$ drives the pedestrian toward its direction with a desired speed.
The pedestrian avoids other neighboring walkers $j=1:N$  as a reaction to $\sum_{j=1}^{N}\vec{f}_{soc,j}$, the summation of their repulsive forces.
Similarly, the pedestrian avoids the boundaries in the vicinity $k=1:B$ due to the pressure from their repelling forces $\sum_{k=1}^{B}\vec{f}_{bnd,k}$.
\par
Traditional SFM replicates the pedestrian walking behavior by assuming centered mass dynamics under $\vec{F}$.
Our mobile robot navigates the sidewalk following the same concept~\cite{Jafari2026-TRO}.
However, due to its nonholonomic constraints, the robot reacts to $\vec{F}$ through its constrained dynamics, unlike the free mass pedestrian.
As a result, the robot's reaction to the virtual force $\vec{F}$ is compatible with its kinematics.
The concept is similar to the work by Gu et al., where a robot dog and a human cooperate to transport goods~\cite{Gu2025}.
Gu et al. suggest a hierarchical algorithm for cooperation between a quadruped robot guided by a physical leash.
We use a virtual force to move through a crowd instead of an actual leash.

\par
Assume $q=[x, y, \theta]^T$ and $z=[v, \omega]^T$, where $x$, $y$, $\theta$ are the robot's center of rotation coordinates, and its heading; $v$ and $\omega$ are its translational and rotational velocities.
The Nonholonomic Mobile Robot (NMR) dynamic mode is ~\cite{Fierro1995}
\begin{equation}
M\Ddot{q}+C\dot{q}=Bu-A^T\lambda_L,
\label{eq:Dynamics}
\end{equation}
where $M$, $C$, and $B$ are the inertia, centrifugal and Coriolis, and input matrices, respectively.
$\lambda_L$ is the Lagrangian multiplier, and $A^T\lambda_L$ represents the forces required to enforce the kinematic constraints; $u=\vec{F}=[F_m, F_n]^T$.
The reduced dynamics of the NMR are~\cite {Jafari2026-TRO}
\begin{equation}
M_r\dot{z}+C_rz=B_ru,
\label{eq:DynamicsFinal}
\end{equation}
with $M_r=J^TMJ$, $C_r=J^TM\dot{J}+J^TCJ$, and $B_r=J^TB$ or 
\begin{equation}
\left\{
\begin{aligned}
    & m\dot{v}-mb\omega^2 = F_m\\
    & (I+mb^2)\dot{\omega}+mb\omega v = bF_n
\end{aligned}
\right. ,
\label{eq:DynamicsSimple}
\end{equation}
where $m$, $I$, and $b$ are the robot's virtual mass, virtual inertia, and virtual torque arm.
\subsection{Projected time-to-collision}
On a straight line, time-to-collision (TTC) is the time remaining until a collision if the robot and the pedestrian keep moving at the same speed.
TTC depends on the distance between the two objects and their approach rate.
Similarly, on a two-dimensional space, Projected TTC (PTTC) is the distance between the two objects divided by their approach rate on the line of sight~\cite{JAFARI2024-2-NATCOM, JAFARI2024-3-SORO, Jafari2023-2-IFAC}. 
Assume $\vec{d}_j$ is the vector connecting the pedestrian to the robot.
Their relative velocity is $\vec{v}_j=\vec{v}_{ped}-\vec{v}$, where $\vec{v}_{ped}$ and $\vec{v}$ are the pedestrian and the robot's absolute velocities.
Then, PTTC, denoted by $t_{c,j}$, is~\cite{Jafari2026-TRO}
\begin{equation}\label{eq:tcj} 
t_{c,j}=\frac{\|\vec{d}_j\|}{\vec{v}_{j\parallel\vec{d}_j}}=\frac{\|\vec{d}_j\|^2}{\vec{d}_j\cdot\vec{v}_j},
\end{equation}
where $\|\vec{d}_j\|$ is the distance between the robot and the pedestrian and $\vec{v}_{j\parallel\vec{d}_j}$ is the approach rate on the line of sight.
Since PTTC correlates with pedestrian subjective comfort in e-scooter-pedestrian interactions~\cite{JAFARI2024-2-NATCOM}, we formulate the introduced forces in various SFM variants using the relative distance (the basic form) and PTTC.
\par
\section{Navigation algorithm variants}\label{sec:theory:main}
Following the pedestrian formulations~\cite{Helbing2000}, we first define the desired force $\vec{f}_{des}$ and the boundary forces $\vec{f}_{bnd,k}$.
In this paper, $\vec{f}_{des}$ and $\vec{f}_{bnd,k}$ are the same among the SFM variants and $\vec{f}_{soc,j}$ changes.
We obtain $\vec{F}$ using \eqref{eq:SFMForce}, calculate the $z=[v,\;\omega]^T$ using \eqref{eq:DynamicsFinal}--\eqref{eq:DynamicsSimple}, and send the $v$ and $\omega$ as commands to the robot base.
Therefore, the robot realizes its desired velocity considering the pedestrians and boundaries while conforming to its kinematic constraints.
Fig.~\ref{fig:SFM_illustration} introduces the forces and their corresponding distance vectors.
\par
The desired force $\vec{f}_{des}$ drives the robot in a certain direction with a certain speed to realize a desired velocity $\vec{v}_{des}$; $m$ is the robot's virtual mass and the relaxation time $\tau_d$ determines the robot's aggressiveness in realizing $\vec{v}_{des}$,
\begin{align}
\vec{f}_{des}=m\frac{\vec{v}_{des}-\vec{v}}{\tau_d}.
\label{eq:f_des}
\end{align}

\par
While most pedestrian models use constant desired velocity, a varying $\vec{v}_{des}$ better fits an autonomous agent in crowded environments~\cite{JAFARI2024-3-SORO, JAFARI2024-SIMPAT},
\begin{align}
\vec{v}_{des}&=v_{max}\;\exp{\left(-\frac{S}{\sigma}\right)}\;\vec{e}_{des},\label{eq:v_des}\\
S&=\sum_{j=1}^{N}\|\vec{f}_{soc,j}\|+\sum_{k=1}^{B}\|\vec{f}_{bnd,k}\|,
\label{eq:S}
\end{align}
where $v_{max}$ is the maximum speed at $S=0$, i.e., no boundary and no pedestrian case.
In case of a very crowded neighborhood, $S=\infty$, $v_{max}$ drops to zero; $\sigma$ tunes the drop rate.
$\vec{e}_{des}$ is a unit vector pointing to the desired direction.
In this paper, we set $v_{max}$=2.0 m/s.
\par
The boundary force $\vec{f}_{bnd,k}$ pushes the robot away from the boundary $k$,
\begin{equation}
\vec{f}_{bnd,k}\left(\Vec{d}_k\right)=\alpha_{b}\exp{\left(-\frac{\|\vec{d}_k\|}{\beta_{b}}\right)}\vec{e}_k,
\label{eq:f_bnd}
\end{equation}
where $\alpha_b$ and $\beta_b$ are the magnitude and range constants.
$\vec{d}_k$ is a distance vector, perpendicular to the boundary $k$ and pointing to the center of rotation with unit vector $\vec{e}_k$.
\par
Original SFM uses the distance vector from the pedestrian $\vec{d}_j$ to describe $\vec{f}_{soc,j}$~\cite{Helbing2000},
\begin{equation}
\vec{f}_{soc,j}\left(\vec{d}_j\right)=\alpha\exp{\left(-\frac{\|\vec{d}_j\|}{\beta}\right)}\vec{e}_j,
\label{eq:f_soc:SFM}
\end{equation}
with $\alpha$ and $\beta$ as the magnitude and range constants.
$\vec{d}_j$ is a distance vector, starting from the pedestrian $j$ and pointing to the center of rotation with unit vector $\vec{e}_j$.
\par
PTTC-based SFM (TSFM) replaces the distance dependency with PTTC dependency. 
TSFM models the robot's reaction to pedestrian $j$ as~\cite{Jafari2026-TRO, JAFARI2024-3-SORO}
\begin{align}
\vec{f}_{soc,j}=\exp{\left(-\frac{t_{c,j}-T_{cr}}{\beta_T}\right)}\vec{e}_{j},
\label{eq:f_soc:TSFM}
\end{align}
where $t_{c,j}$ is the projected time-to-collision to pedestrian $j$ and $\vec{e}_j=\vec{d}_j/\|\vec{d}_j\|$.
\par
\begin{figure}
\centering
\includegraphics[width=0.45\textwidth]{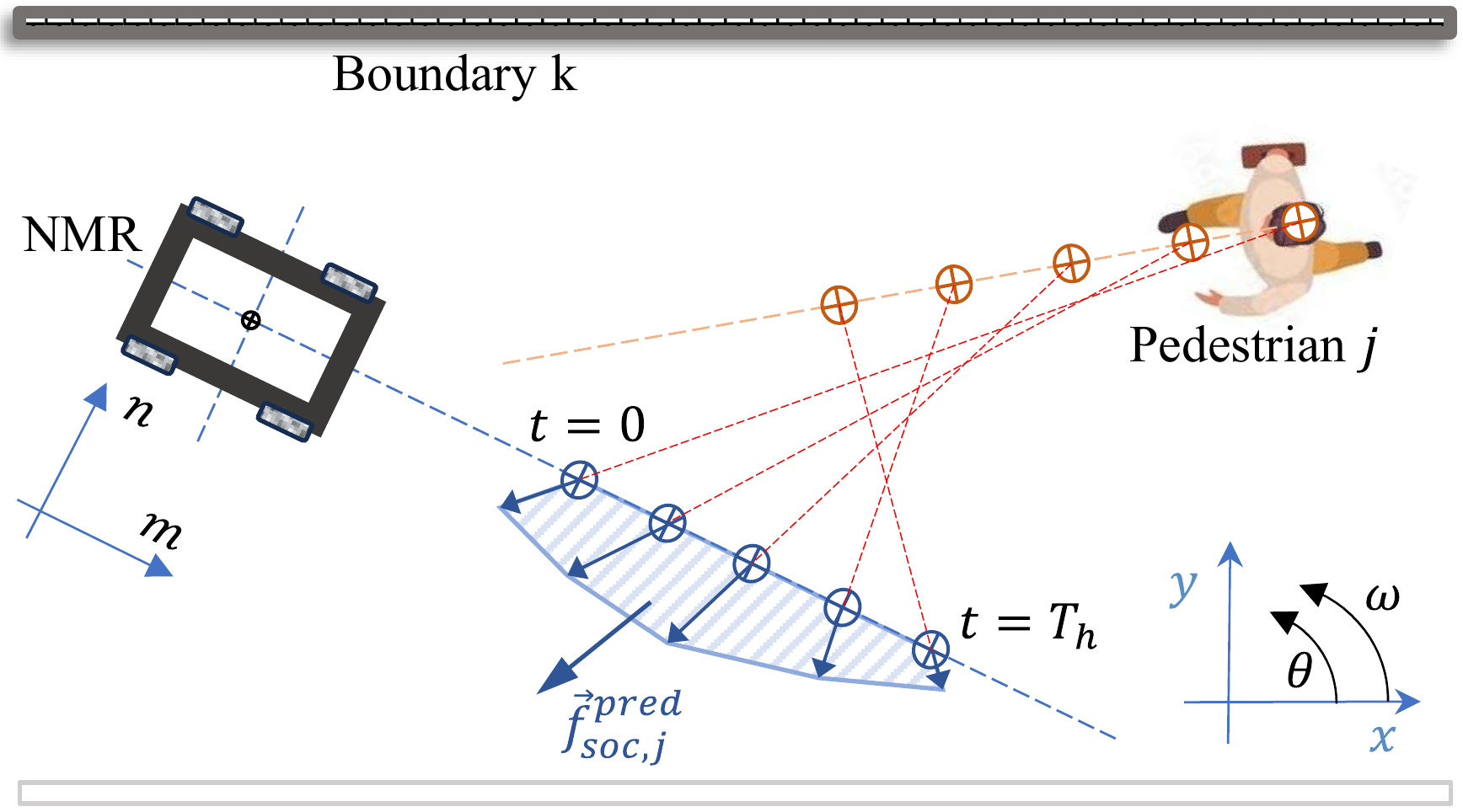}
\caption{The social force vector field over the next $T_h$ seconds.
The predictive social force variants, PSFM and PTSFM, average $\vec{f}_{soc,j}$ over the time horizon $[0,\;T_h]$.
Because of the assumption of constant velocities, the relative velocity does not change. The relative position vector is~\eqref{eq:predicted_d_j}.
}
\label{fig:Prediction_illustration}
\end{figure}

In this paper, we incorporate the predicted evolution of $\vec{f}_{soc,j}$ into the mobile robot's SFM and TSFM and call them Predictive SFM (PSFM) and Predictive TSFM (PTSFM), respectively.
We assume constant velocities for a certain time horizon $T_h$.
Therefore,
\begin{align}
\vec{d}_j\left(\tau\right)=\vec{d}_j\left(0\right)+\vec{v}_j\tau,
\label{eq:predicted_d_j}
\end{align}
where $\tau$ is a prediction instance of the prediction horizon, $\tau\in[0,\;T_h]$, and corresponds to $\vec{f}_{soc,j}(\tau)$ determined by~\eqref{eq:f_soc:SFM} and~\eqref{eq:f_soc:TSFM}.
We average $\vec{f}_{soc,j}(\tau)$ over the span $[0,\;T_h]$ similar to e-scooters~\cite{JAFARI2024-3-SORO}. 
Thus, in PSFM and PTSFM, 
\begin{equation}
    \vec{f}_{soc,j}^{pred}=\frac{1}{T_h}\int_{0}^{T_h} \vec{f}_{soc,j}\left(\tau\right) \;d\tau.
\label{eq:f_soc:PSFMPTSFM}
\end{equation}
where $\vec{f}_{soc,j}^{pred}$ is the predictive social force and replaces $\vec{f}_{soc,j}$ in SFM and TSFM.
Fig.~\ref{fig:Prediction_illustration} illustrates the predicted vector field in the prediction horizon.
In this paper, all the future moments are equally important.
An alternative is a decaying weight factor, for example,
\begin{equation}
    \vec{f}_{soc,j}^{pred}=\int_{0}^{\infty} e^{-\tau}\vec{f}_{soc,j}\left(\tau\right) \;d\tau.
\label{eq:f_soc_decay:PSFMPTSFM}
\end{equation}
\par
Overall, the studied navigation algorithms in this paper are SFM, TSFM, PSFM, and PTSFM.
SFM is a distance-based method, TSFM is a time-to-collision-based algorithm, PSFM is SFM with prediction, and PTSFM is TSFM with prediction.
We experimentally study the method's performance on pedestrians' objective and subjective safety.
\par
\section{Experiments}\label{sec:exp:main}
This section contains the experiment setup, the selected parameters, and the design of the experiments.
Section~\ref{sec:exp:setup} focuses on the hardware and the software to implement and perform the navigation.
Section~\ref{sec:exp:doe} describes the trials, the critical variables, the scenario, and some details about the participants.
National Cheng Kung University (NCKU) ethical board approved all the procedures, and all participants provided written consents.
\subsection{Setup}\label{sec:exp:setup}
\begin{figure*}
    \centering
    \subfloat[]{
        \includegraphics[width=0.232\textwidth]{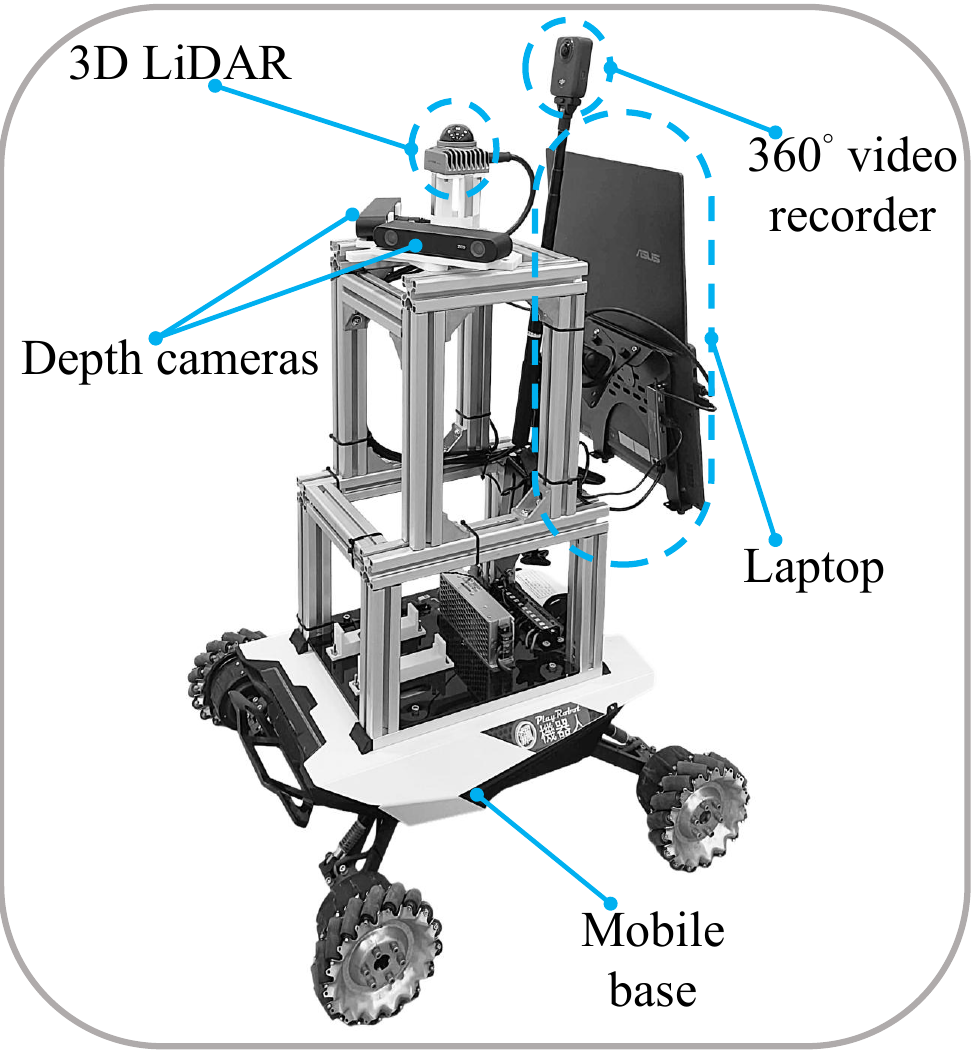}
    }
    \subfloat[]{
        \includegraphics[width=0.40\textwidth]{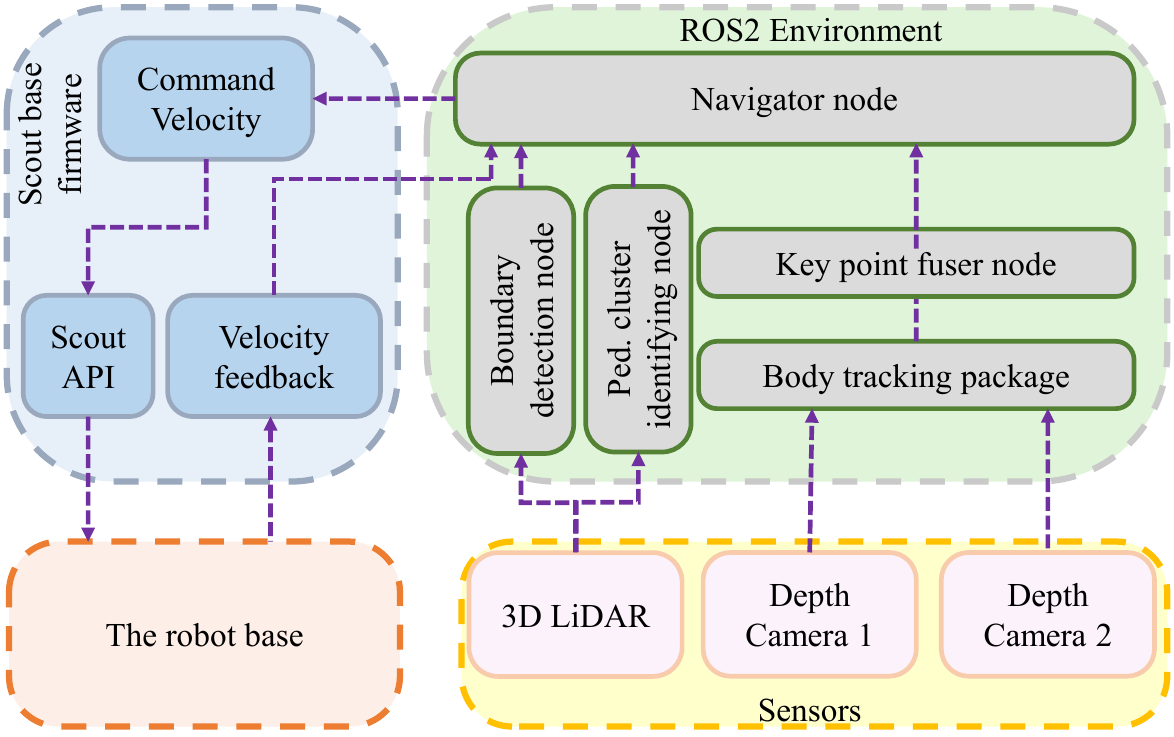}
    }
    \subfloat[]{
        \includegraphics[width=0.177\textwidth]{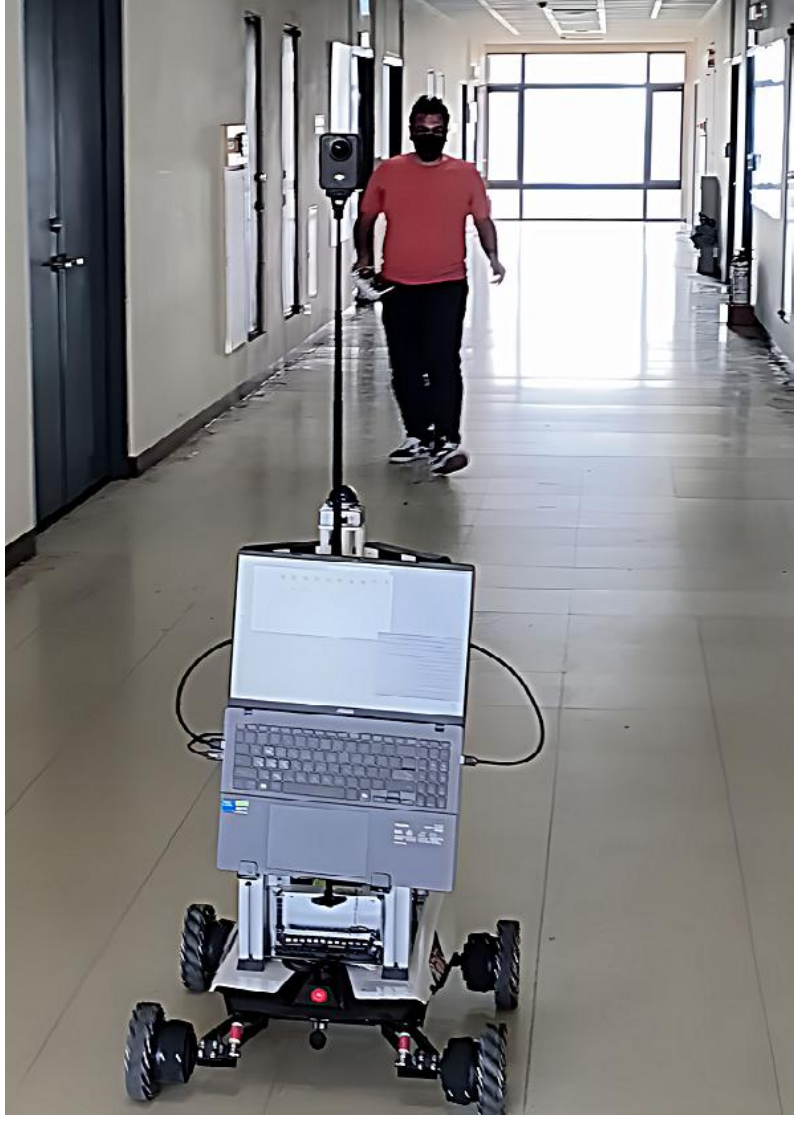}
    }
    \caption{The experiment setup. 
    (a) An Agilex Scout Mini is the mobile base. 
    Two Zed2i depth cameras see the walkers. 
    A Livox Mid360 detects the boundaries. 
    An ASUS V16 laptop is the processor.
    A 360-degree camera records the trial.
    (b) The block diagram of the whole system:
    The sensors, including the LiDAR and the camera, measure the required inputs.
    The ROS2 navigator node gets the topics, calculates the commands, and publishes them to the base.
    The base realizes the commands and publishes the feedback.
    (c) The robot faces a pedestrian in a trial.}
    \label{fig:ExpSetup}
\end{figure*}
\begin{table}
\renewcommand\arraystretch{1.0}
\centering
\caption{SFM/PSFM and TSFM/PTSFM parameters used in experiments. 
The parameters are selected based on the previous experience~\cite{Jafari2026-TRO} and further experimental trial and error.}
\begin{tabular}{||b{0.16\textwidth}|
                >{\centering\arraybackslash}b{0.07\textwidth}|
                >{\centering\arraybackslash}b{0.04\textwidth}|
                >{\centering\arraybackslash}b{0.05\textwidth}||}
\hline
\textbf{Name} & \textbf{Symbol (units)} & \textbf{SFM PSFM} & \textbf{TSFM PTSFM} \\ 
\hline\hline
\multicolumn{4}{||l||}{\textbf{Robot dynamics constants}} \\
\hline
Mass & $m~(kg)$ & 20 & 20 \\
\hline
Inertia & $I~(kg.m^2)$ & 1.0 & 1.0 \\ 
\hline
Torque arm & $b~(m)$ & 0.5 & 0.5 \\ 
\hline
\multicolumn{4}{||l||}{\textbf{Pedestrian force constants}} \\
\hline
Relaxation time & $\tau_d~(s)$ & 0.3 & 0.3 \\ 
\hline
Force magnitude constant & $\alpha~(N)$ & 183.5 & -- \\
\hline
Force range & $\beta~(m)$ & 1.9 & -- \\ 
\hline
Desired speed adjuster & $\sigma~(N)$ & 140 & 140 \\
\hline
PTTC shift & $T_{cr}~(s)$ & -- & 2.5 \\ 
\hline
Force range & $\beta_T~(s)$ & -- & 0.2 \\ 
\hline
Prediction horizon & $T_h~(s)$ & 0\;/\;2.0 & 0\;/\;2.0  \\ 
\hline
\multicolumn{4}{||l||}{\textbf{Boundary force constants}} \\ 
\hline
Force magnitude & $\alpha_b~(N)$ & 300 & 300 \\
\hline
Force range & $\beta_b~(m)$ & 0.5 & 0.5 \\
\hline
\end{tabular}
\label{tab:setpar}
\end{table}
Fig.~\ref{fig:ExpSetup}(a) is the physical setup used for the experiments.
Briefly, we installed depth cameras on the platform to detect the pedestrians and measure their relative distance and velocity with respect to the robot's local frame.
The cameras track the pedestrian body keypoints from the RGB+D data.
A 3D LiDAR collects the point cloud of the robot's surroundings for boundary detection.
Specifically, it detects walls by running a RANSAC algorithm on points higher than pedestrian height and lower than the ceiling.
A laptop receives the information from the sensors, processes the algorithm, and sends the command signals to the robot base using ROS2 Humble middleware. 
\par
The cameras, a pair of StereoLabs Zed2is, are symmetrically fixed at the top level of the robot frame with $\pm33^\circ$ yaw angles.
They collect the depth frames and detect the body's skeletal features. 
The Livox LiDAR Mid-360 is installed at the robot's local frame origin, center of rotation. 
It collects the 360-degree point cloud of the experiment hall.
The mobile base is Agilex Scout Mini.
The mobile base has four Mecanum wheels. 
However, since the lateral velocity is set to zero, it acts as a differential drive mobile robot.
A laptop, ASUS V16 with RTX 3050 GPU running Ubuntu 22.04, acts as the central processor.
A DJI Osmo camera records the trials independently.
\par
Fig.~\ref{fig:ExpSetup}(b) shows the algorithm implementation chart.
A package extracts the pedestrians' skeletal features from the RGB-D frames for each camera.
A node transforms the camera readings into the robot's local frame. 
Since each camera's FOV is 114 degrees and they have about $66^\circ$ yaw difference, another node unifies the key points of the readings, removes duplicate detections from the overlapping areas, and publishes the pedestrians' states to the navigator node.
Moreover, a node extracts the distance to the walls from the LiDAR point cloud.
Another node locates the pedestrian cluster in the cameras' blind spots.
The mobile robot base publishes the translational and rotational velocities to the navigator.
\par
The navigator node receives the pedestrian relative position and velocity topics, the walls' position topics, the robot's instantaneous translational and angular speeds, and its heading.
It calculates the forces and sends the velocity commands to the robot base node; the robot's lateral velocity is set to zero as a kinematic constraint.
The base node ensures that the commands are realized using an internal controller.
ROS2 rqt package identifies the following communication rates: 10 Hz for the pedestrian states collected by the cameras, 15 Hz for the wall distance by the LiDAR, 12 Hz for the pedestrian position detected by the LiDAR, 10 Hz for feedback from the robot base, and 50 Hz for the commands sent to the base.
\par
Regarding the required parameters for the navigation, we use the previously reported values obtained by simulation~\cite{Jafari2026-TRO}.
However, we further refine them experimentally by trial and error.
Table~\ref{tab:setpar} presents the assumed robot dynamic parameters, the pedestrian force constants, and the boundary force parameters.
%%%%%%%%%%%%%%%%%%%%%%%%%%%%%%%%%%%%%%%%%%%%%%%%%%%%%%%%%%%%%%%%%%%%%%%%%%%%%%%%%%%%%%%%%%%%%%%%%%%%%%%%%%%%%%%%%%%%%%%%%%%%%%%%%
%%%%%%%%%%%%%%%%%%%%%%%%%%%%%%%%%%%%%%%%%%%%%%%%%%%%%%%%%%%%%%%%%%%%%%%%%%%%%%%%%%%%%%%%%%%%%%%%%%%%%%%%%%%%%%%%%%%%%%%%%%%%%%%%%
%%%%%%%%%%%%%%%%%%%%%%%%%%%%%%%%%%%%%%%%%%%%%%%%%%%%%%%%%%%%%%%%%%%%%%%%%%%%%%%%%%%%%%%%%%%%%%%%%%%%%%%%%%%%%%%%%%%%%%%%%%%%%%%%%
%%%%%%%%%%%%%%%%%%%%%%%%%%%%%%%%%%%%%%%%%%%%%%%%%%%%%%%%%%%%%%%%%%%%%%%%%%%%%%%%%%%%%%%%%%%%%%%%%%%%%%%%%%%%%%%%%%%%%%%%%%%%%%%%%
\subsection{Design of the experiments}\label{sec:exp:doe}
In this section, we present the experiment design and the participant biases and demographics.
Fig.~\ref{fig:ExpSetup}(c) is a typical trial where the robot and the pedestrian face each other and pass by.
In all trials, the mobile robot moves toward a pedestrian from one end of the experiment corridor, and the pedestrian walks toward the robot from the other end.
The hall width is about 3.2 meters, and the pedestrian and the robot start at the same time from an approximately 30-meter distance.
The instruction to the pedestrian is to walk comfortably and react to the robot as if it were a bicycle on a sidewalk to reproduce a bilateral avoidance interaction.
\par
We perform the trials on the National Cheng Kung University (NCKU) campus and recruit the participants through a board announcement in the mechanical engineering department.
The ten male participants are 22.9 years old on average, with a standard deviation of 1.5.
Six out of ten participants reported previous interactions with mobile robots.
Because of the volunteer sampling, we asked the participants to rate their opinion about the safety of walking around mobile robots.
On a 5-point Likert scale, they reported a high level of trust, 4.0 on average, with a standard deviation of 0.5.
\par
Each participant performs twenty trials, five per method (SFM, TSFM, PSFM, and PTSFM), resulting in 200 trials.
The LiDAR detections of one trial were discarded due to recording problems.
During the trials, we record the PTTC, the robot's translational and rotational speeds, its heading, the relative distance and velocity measured by the cameras, the distance to the walls, and the pedestrian's relative position measured by the LiDAR.
After the trials, the participants were asked to rate the interaction considering four aspects: the general comfort, the robot movement smoothness, the appropriateness of the distance, and whether the speed felt comfortable.
\par
\section{Results}\label{sec:res:main}
This section presents the collected data from the trials and discusses the statistical significance of the results.
First, we discuss the qualitative performance of the navigation methods during the trials using the Supplementary video.
Next, we focus on the objective safety of the pedestrians evaluated by PTTC, the average speed in the pedestrians' vicinity, the minimum robot-pedestrian distance, the lateral distance, and the maximum curvature. All these variables are collected/calculated using the onboard sensors.
Then, we study the pedestrian subjective safety using the responses in questionnaires.
Interaction comfort rating, movement smoothness rating, distance comfort rating, and speed comfort rating are the indicators.
A limitation note concludes the section.
\par
\subsection{Video analysis}
The supplementary video contains two performance tests: a free run using PTSFM (overtaking scenario) and a non-cooperative pedestrian, and four controlled trials with a cooperative pedestrian, each presenting a navigation algorithm (facing scenario).
The supplementary video shows the performance of PTSFM interacting with a non-cooperative pedestrian during an overtaking scenario.
The robot tries to realize its desired velocity (2.0 m/s).
Since the pedestrian walks in the same direction at a lower speed, the robot tries to overtake him.
In the video, the pedestrian challenges the robot's decisions multiple times, and the robot successfully reacts to the pedestrian in the complex overtaking scenarios.
Compared to similar scenarios with learning methods like~\cite{Sen2025}, the robot's reactions to unseen scenarios with real-time speed highlight the advantages of SFM variants.
\par
Moreover, qualitative examination of the videos shows that the predictive methods (PSFM and PTSFM) are quite visually similar to their non-predictive versions (SFM and TSFM).
The small differences between the groups are not apparent in the individual trials. 
Nevertheless, they are detectable across multiple trials.
\par
However, PTTC-based methods (TSFM and PTSFM) are clearly distinguishable from the basic methods (SFM and PSFM).
A distinctive feature is a visibly earlier reaction to the pedestrian, i.e., speed reduction at a longer distance.
Another feature is that TSFM and PTSFM nearly stop just before reaching the pedestrian.
During the tests, some participants interpreted this stop as a sign of hesitation and uncertainty in the robot's behavior and reported higher discomfort levels.
This is an example of an objectively safe behavior interpreted as subjectively unsafe.
\par
\subsection{Objective safety}\label{sec:res:objsaf}
\begin{figure*}
\centering
\subfloat[]{\includegraphics[width=0.19\textwidth]{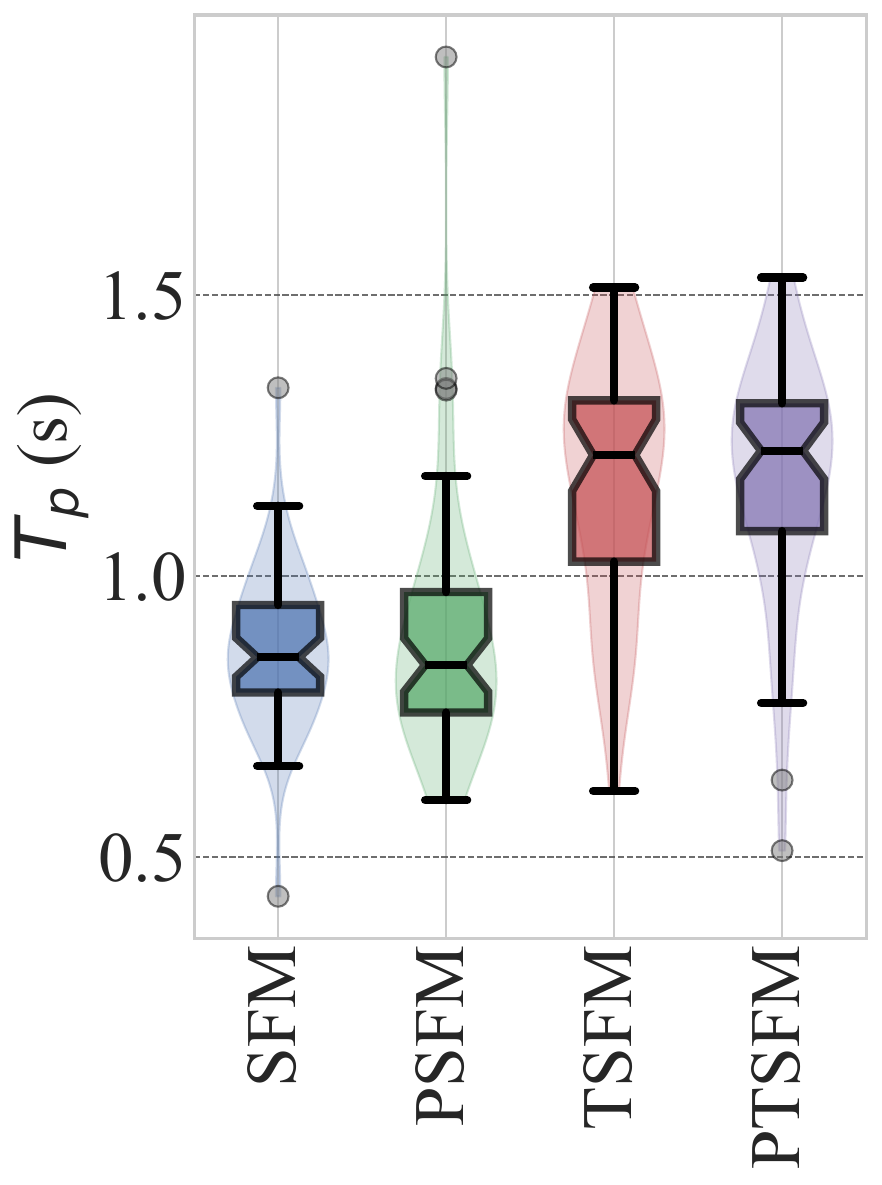}}
\subfloat[]{\includegraphics[width=0.19\textwidth]{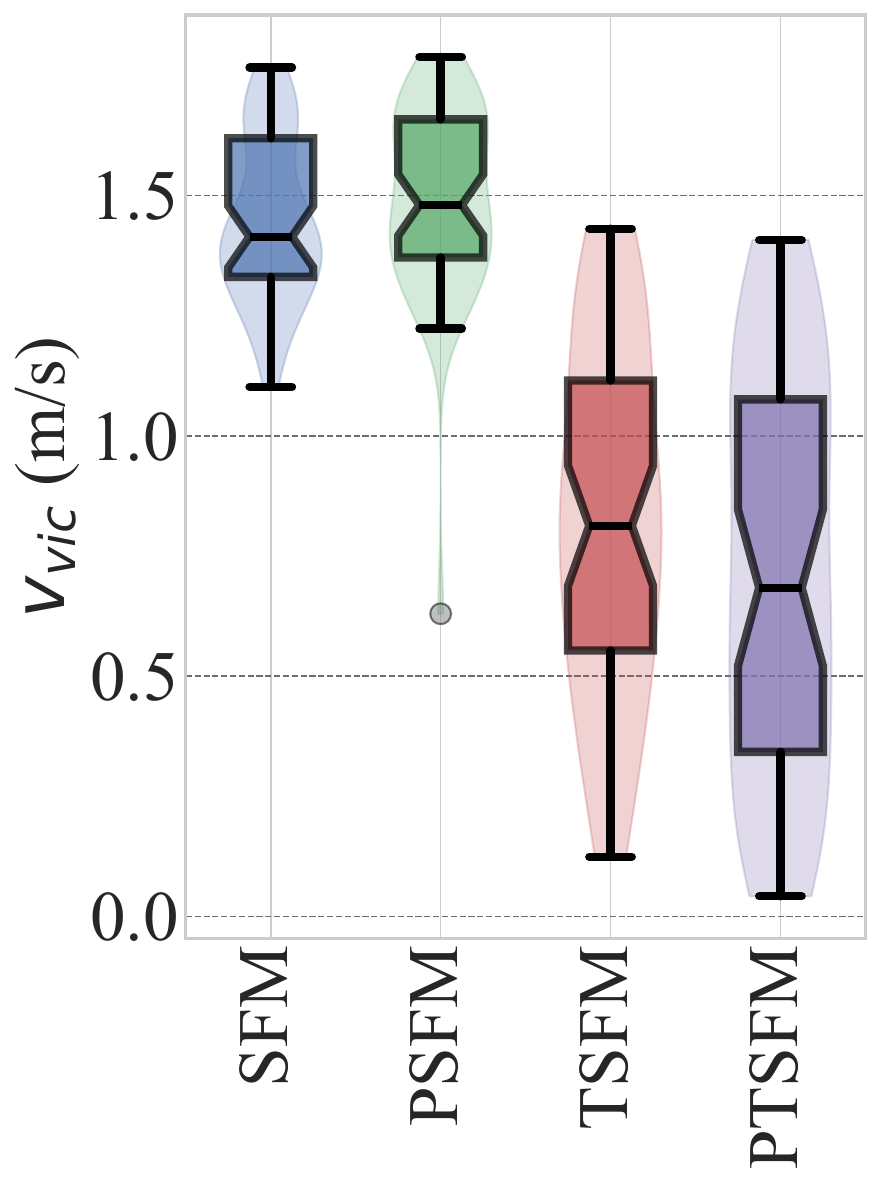}}
\subfloat[]{\includegraphics[width=0.19\textwidth]{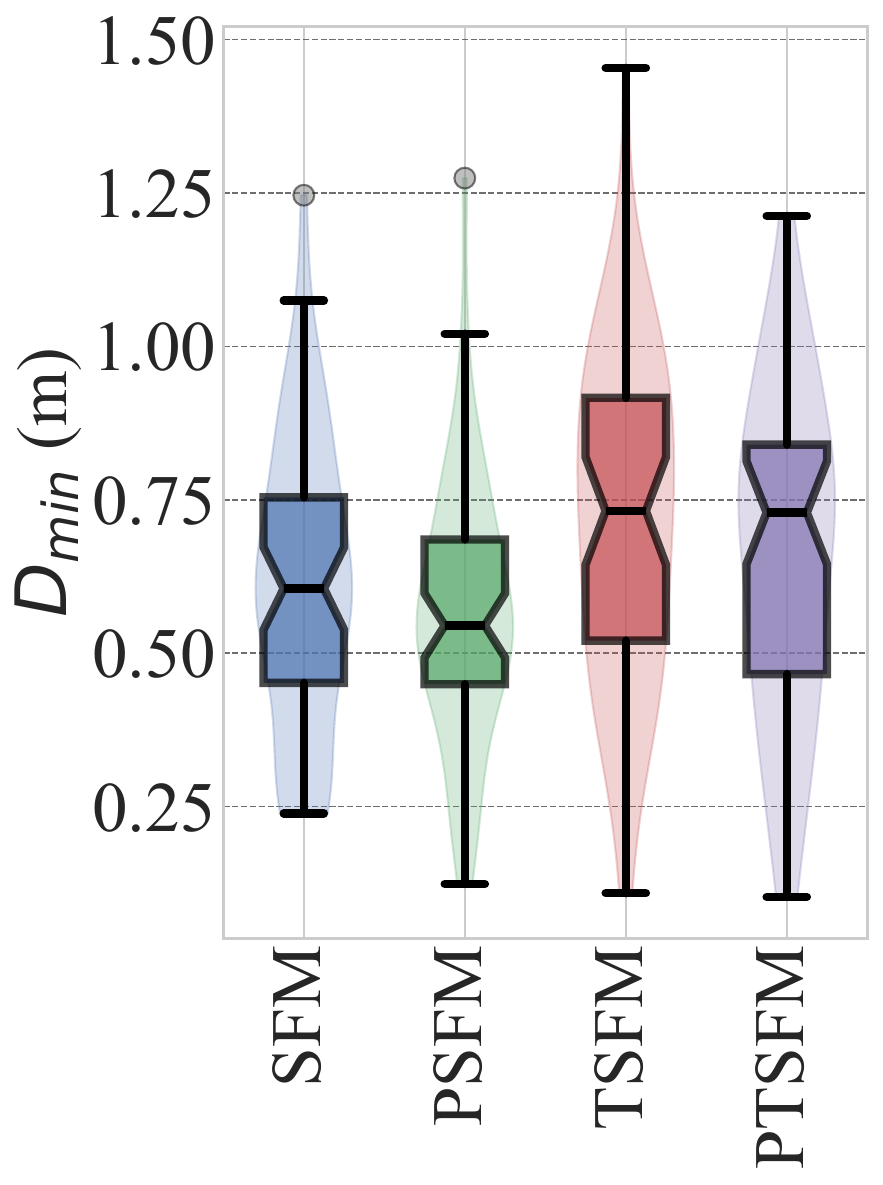}}
\subfloat[]{\includegraphics[width=0.19\textwidth]{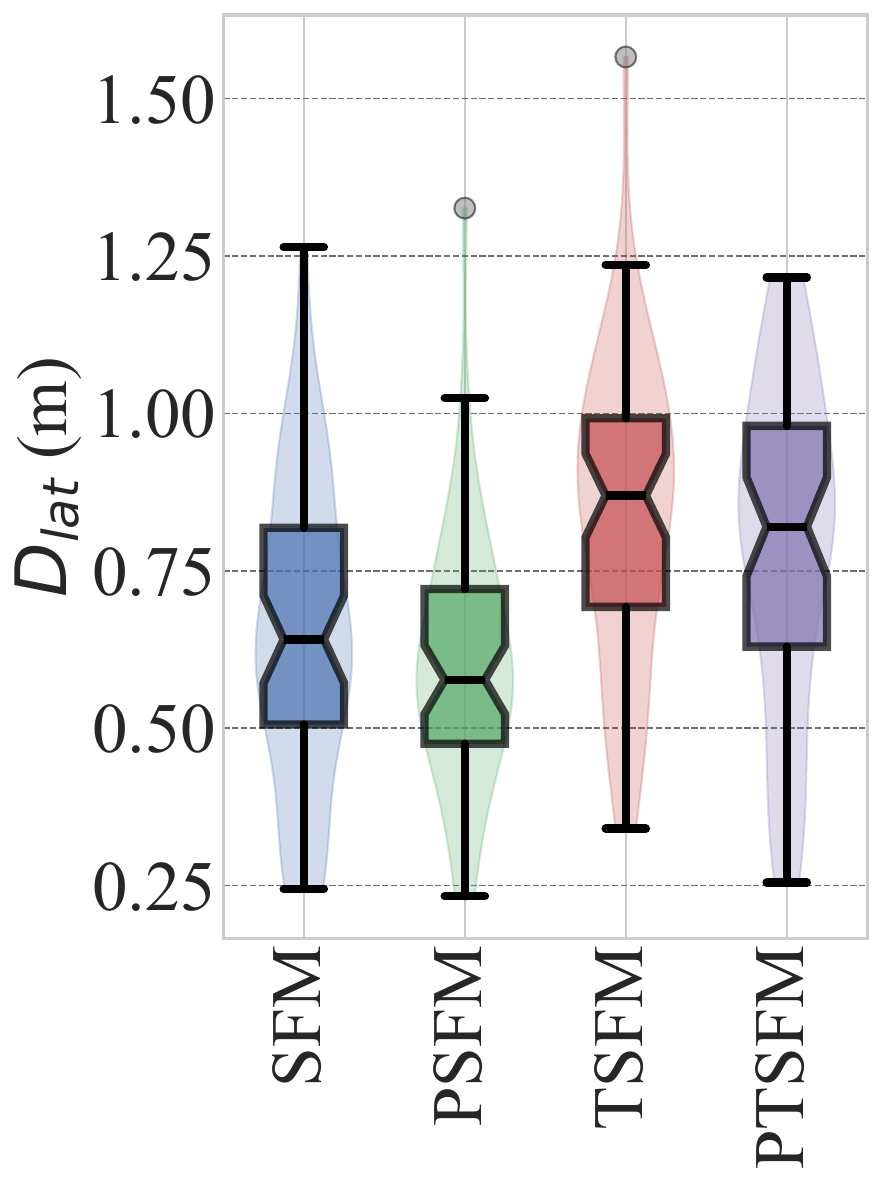}}
\subfloat[]{\includegraphics[width=0.19\textwidth]{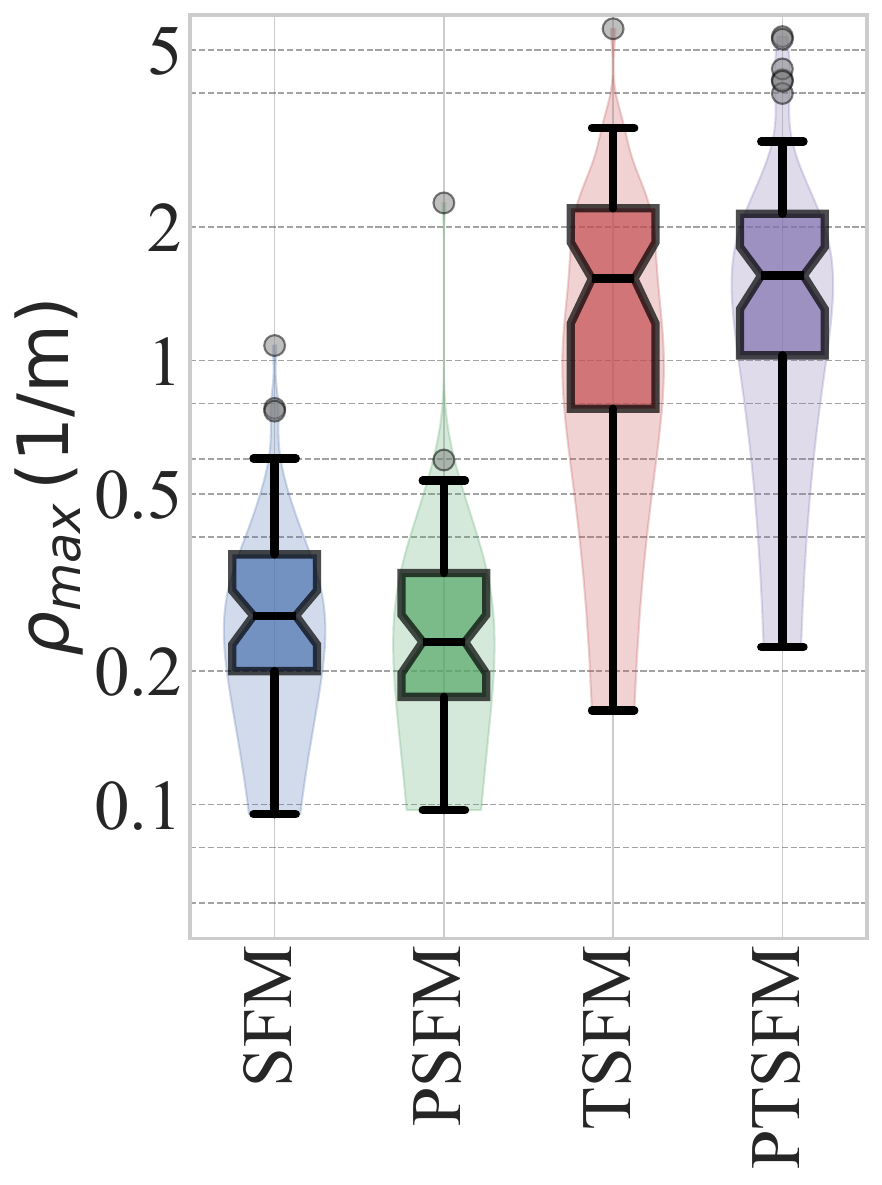}}
\caption{The box/violin plots of the collected data using the recorded trajectories trials. 
Each plot contains four navigation groups: SFM, TSFM, PSFM, and PTSFM. All boxes have 50 data points, except for PTSFM in (c) and (d), which have 49.
(a) $T_p$: the minimum of PTTC $t_{c,j}$ during the trials.
(b) $v_{vic}$: the average speed in the vicinity of the pedestrian.
(c) $D_{min}$: the minimum distance to the pedestrian.
(d) $D_{lat}$: the lateral distance at passing moment.
(e) $\rho_{max}$: the maximum trajectory curvaure.
}
\label{fig:objboxes}
\end{figure*}
The kinematic variables used in this section are defined as follows:
\begin{itemize}
    \item \textbf{Minimum of predicted time-to-collision ($T_p$)} is the minimum of PTTC during each trial. We use the recorded $t_{c,j}$ as a safety metric.
    \item \textbf{Average speed in pedestrian vicinity ($v_{vic}$)} is calculated by finding the mean value of $v$ from the moment the robot enters a 5-meter distance of the pedestrian until it passes by.
    \item \textbf{Minimum distance to the pedestrian ($D_{min}$)} is the closest the robot gets to the walker during each trial.
    \item \textbf{Lateral distance to the pedestrian ($D_{lat}$)} is the distance between the robot and the pedestrian when they are side by side.
    \item \textbf{Maximum trajectory curvature ($\rho_{max}$)} corresponds to the smallest arc fitted to the robot's trajectory using a moving window of length 20 cm.
\end{itemize}
\par
Fig.~\ref{fig:objboxes} summarizes the recorded variables using the box/violin plots across the four methods.
In terms of $T_p$, TSFM and PTSFM outperform SFM and PSFM in Fig.~\ref{fig:objboxes}(a). 
The plot indicates a significant improvement in objective safety when PTTC is incorporated. 
However, the prediction gain, either PSFM over SFM or PTSFM over TSFM, is marginal.
Regarding average speed in Fig.~\ref{fig:objboxes}(b), SFM and PSFM are noticeably faster, but the prediction integration increases the average speed for SFM and decreases it for TSFM.
\par
Fig.~\ref{fig:objboxes}(c) and (d) are closely correlated because during the trials, there is not enough time difference between the moments they happen.
Therefore, usually higher $D_{min}$s result in longer $D_{lat}$s.
Considering the trajectory curvature, Fig.~\ref{fig:objboxes}(e) shows a significant difference between the basic methods and the PTTC-based methods.
The prediction integration lowers the $\rho_{max}$ for SFM.
However, it does not really change the curvature for TSFM.
Overall, the results indicate that PTTC integration improves the objective safety, aligning with the results in~\cite{Jafari2026-TRO}.
In contrast, the prediction has a limited effect on the objective safety metrics.
\par
\subsection{Subjective safety}\label{sec:res:subsaf}
\begin{figure*}
\centering
\subfloat[]{\includegraphics[width=0.19\textwidth]{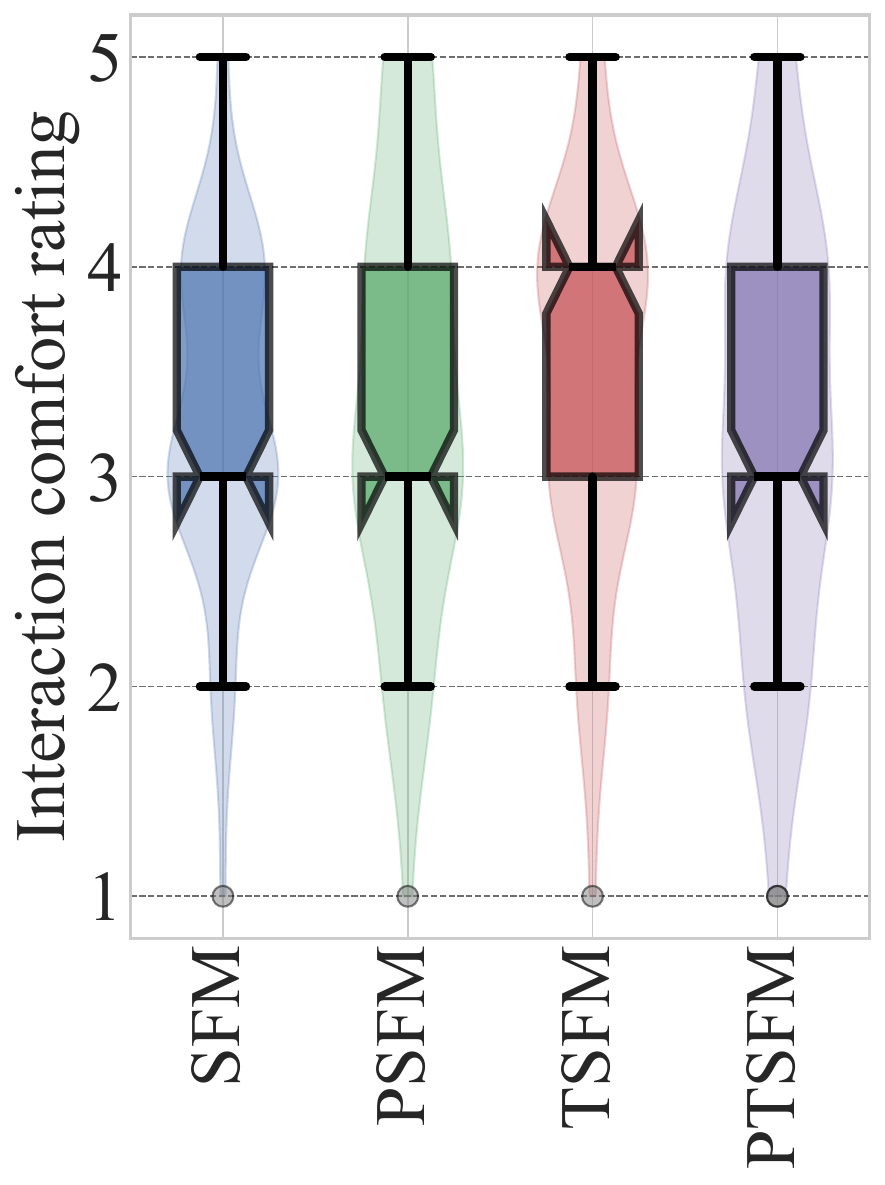}}\hfill
\subfloat[]{\includegraphics[width=0.19\textwidth]{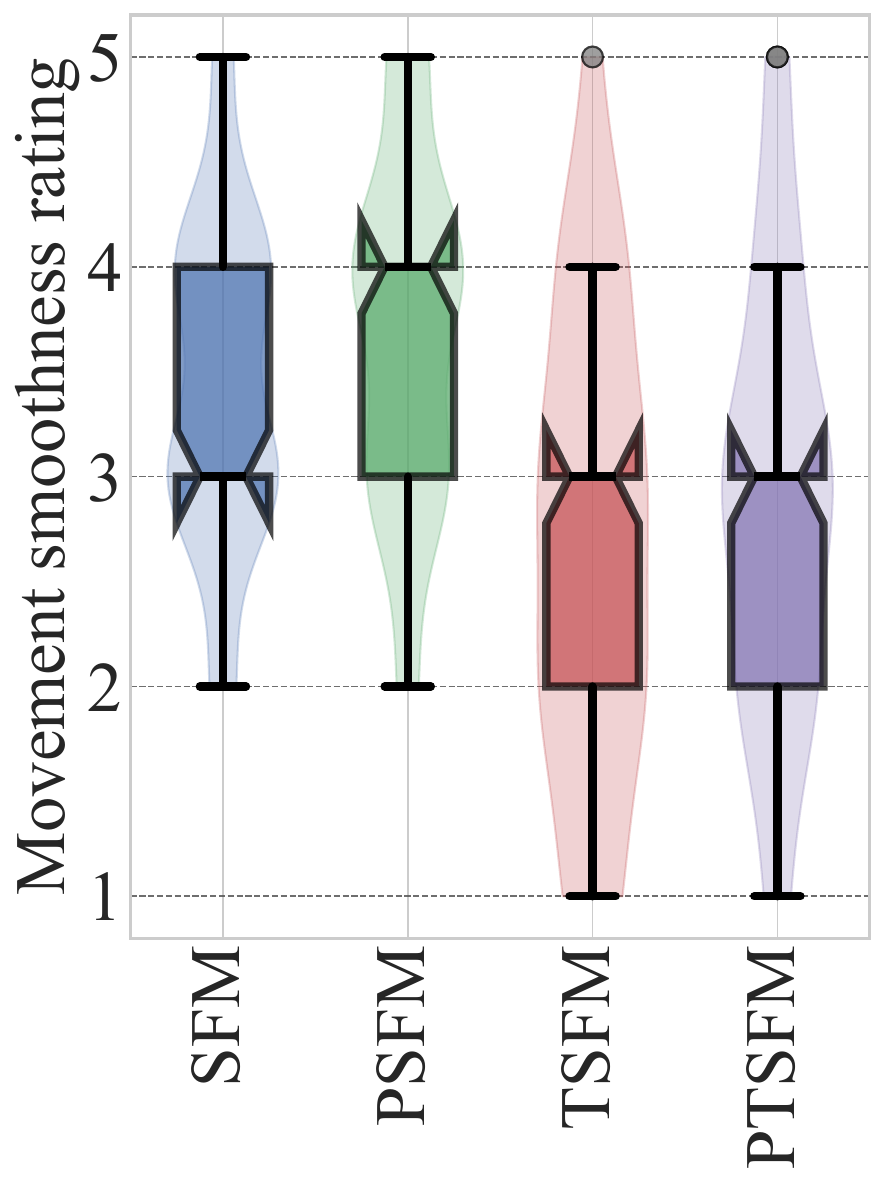}}\hfill
\subfloat[]{\includegraphics[width=0.19\textwidth]{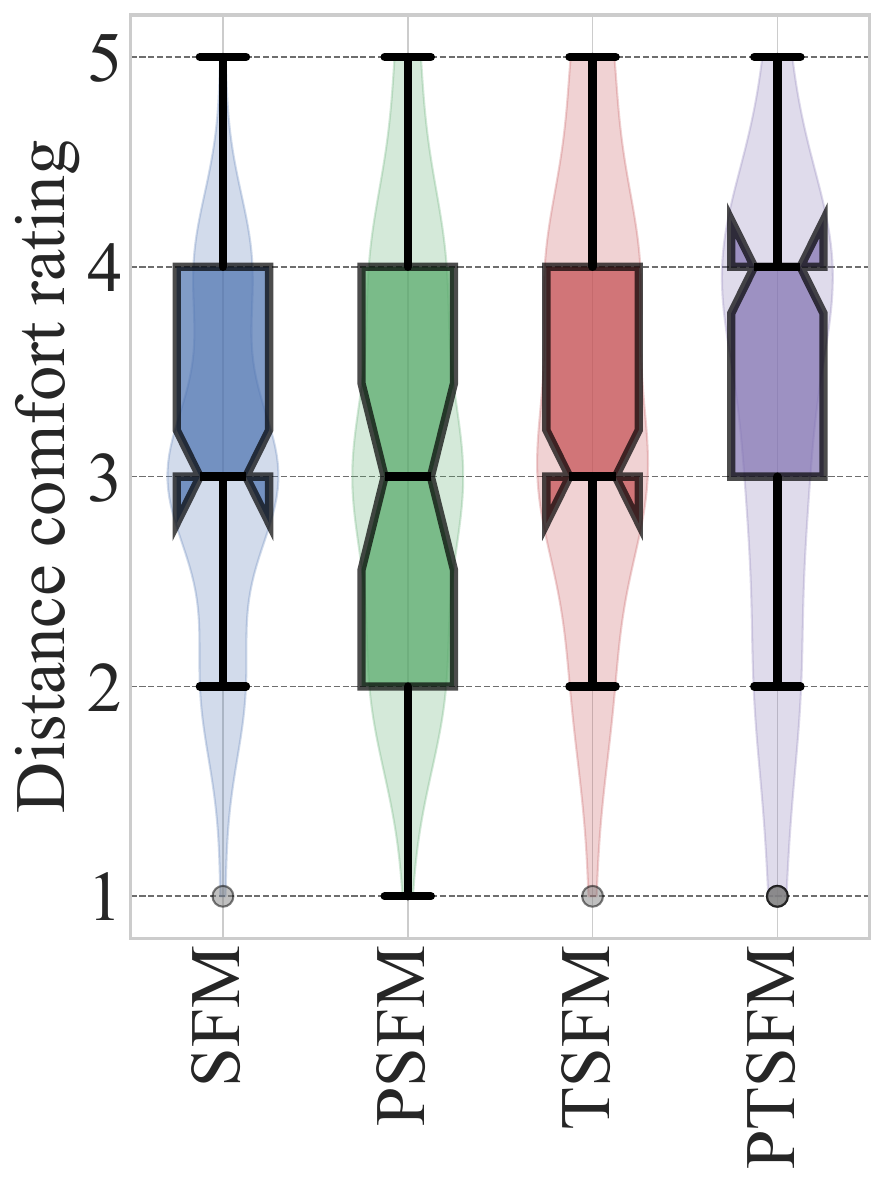}}\hfill
\subfloat[]{\includegraphics[width=0.19\textwidth]{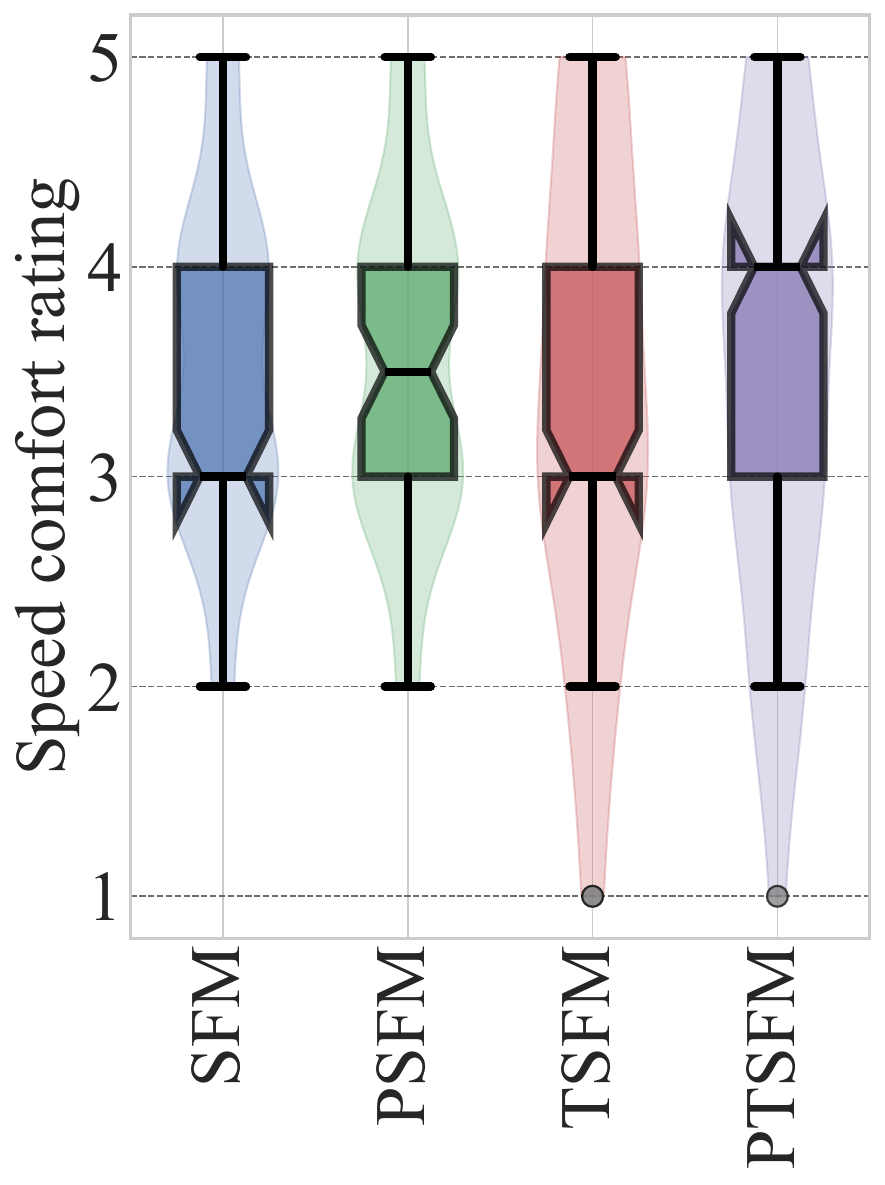}}
\caption{Survey results about pedestrians' opinions on the interaction qualitative features on a 5-point Likert scale.
Each plot contains four navigation groups: SFM, TSFM, PSFM, and PTSFM. All boxes have 50 data points (a total of 200 per plot).
(a) Interaction comfort rating: the participants' opinion about the perceived comfort during the trial.
(b) Movement smoothness rating: the walker's judgment about the smoothness of the robot's trajectory during the test.
(c) Distance comfort rating: whether the pedestrian is comfortable with the distance maintained by the robot.
(d) Speed comfort rating: whether the pedestrian is uncomfortable with the robot's speed.
}
\label{fig:subboxes}
\end{figure*}
Survey responses benchmark SFM, PSFM, TSFM, and PTSFM in this section.
Fig.~\ref{fig:subboxes}(a) is a box/violin plot of interaction comfort rating, marking the participants' opinion about the perceived comfort during the trial.
Fig.~\ref{fig:subboxes}(b) quantifies the walker's judgment about the smoothness of the robot's trajectory during the test.
Fig.~\ref{fig:subboxes}(c) shows whether the pedestrian is comfortable, specifically with the distance maintained by the robot.
Fig.~\ref{fig:subboxes}(d) focuses on whether the pedestrian is uncomfortable with the robot's speed.
\par
Regarding the survey reliability, a Cronbach's test reports an $\alpha$ of 0.82 and shows good internal consistency.
Pairwise correlations are all positive and moderate to strong (r=0.38--0.66), which indicates the factors are related and non-redundant.
In addition, we perform pairwise nonparametric comparisons between SFM and PSFM, and between TSFM and PTSFM, using the Mann-Whitney U test.
Table~\ref{tab:MannW} presents the results with rank-biserial correlation and p-values.
\par
Fig.~\ref{fig:subboxes}(a) highlights that TSFM has the highest comfort rating among the methods, with a median of 4, while others show a median of 3.
However, the violin plot reveals very similar distributions. 
Consistently, the Mann-Whitney tests report no statistically significant difference between SFM and PSFM and between TSFM and PTSFM with p-values of 0.67 and 0.29, respectively; the effect size is negligible($|r_{rb}|<0.12$).
\par
Fig.~\ref{fig:subboxes}(b) indicates that the pedestrians perceived lower movement smoothness during TSFM and PTSFM trials, which is compatible with the recorded higher maximum curvature in Fig.~\ref{fig:objboxes}(e).
In addition, the participants notice the curvature reduction due to the prediction integration, visible in Fig.~\ref{fig:objboxes}(e), and report smoother movement for PSFM compared to SFM (p-value=0.13).
\par
Fig.~\ref{fig:subboxes}(c) shows mostly similar distributions. 
The PTSFM's higher median shows that some participants believe the distance behavior is improved.
For the prediction effect, Table~\ref{tab:MannW} reports the absence of meaningful contributions as the p-values are above 0.8 and the effect sizes are almost zero.
Regarding the speed behavior, the participants believe the predictions have meaningful contributions.
Despite the visible differences and descriptive trends, no statistically significant difference is present.
\par
Overall, the prediction integration weakly improves the opinion of the walkers about the distance and speed behavior of the robot.
\begin{table}[t]
\centering
\caption{Mann--Whitney rank-biserial correlations and two-sided p-values for subjective metrics. 
Each row corresponds to a subplot in Fig.~\ref{fig:subboxes}. 
}
\begin{tabular}{||l|c|c|c|c||}
\hline
\multirow{2}{*}{Metric} & \multicolumn{2}{c|}{SFM vs PSFM} & \multicolumn{2}{c||}{TSFM vs PTSFM} \\
\cline{2-5}
 & $r_{rb}$ & p-value & $r_{rb}$ & p-value \\
\hline
\hline
Interaction comfort rating & 0.05 & 0.67 & -0.12 & 0.29 \\
\hline
Movement smoothness rating & 0.16 & 0.14 & 0.11 & 0.34 \\
\hline
Distance comfort rating & 0.020 & 0.85 & -0.02 & 0.86 \\
\hline
Speed comfort rating & 0.02 & 0.89 & 0.03 & 0.79 \\
\hline
\end{tabular}
\label{tab:MannW}
\end{table}
%%%%%%%%%%%%%%%%%%%%%%%%%%%%%%%%%%%%%%%%%%%%%%%%%%%%%%%%%%%%%%%%%%%%%%%%%%%%%%%%%%%%%%%%%%%%%%%%%%%%%%%%%%%%%%%%%%%%%%%%%%%%%%%%%
%%%%%%%%%%%%%%%%%%%%%%%%%%%%%%%%%%%%%%%%%%%%%%%%%%%%%%%%%%%%%%%%%%%%%%%%%%%%%%%%%%%%%%%%%%%%%%%%%%%%%%%%%%%%%%%%%%%%%%%%%%%%%%%%%
%%%%%%%%%%%%%%%%%%%%%%%%%%%%%%%%%%%%%%%%%%%%%%%%%%%%%%%%%%%%%%%%%%%%%%%%%%%%%%%%%%%%%%%%%%%%%%%%%%%%%%%%%%%%%%%%%%%%%%%%%%%%%%%%%
%%%%%%%%%%%%%%%%%%%%%%%%%%%%%%%%%%%%%%%%%%%%%%%%%%%%%%%%%%%%%%%%%%%%%%%%%%%%%%%%%%%%%%%%%%%%%%%%%%%%%%%%%%%%%%%%%%%%%%%%%%%%%%%%%
\subsection{Limitations}\label{sec:res:lim}
A few caveats limit the generalizability of the results:
\begin{itemize}
    \item The participant pool is homogeneous and biased. 
          Our volunteers are young male recruits from the mechanical engineering department with high prior trust in mobile robots.
    \item We only examined indoor and single pedestrian trials in facing encounter scenarios. 
          The controlled experimental environment differs from actual public spaces.
    \item We fixed the prediction horizon at $T_h=2$~s by observing robots' behavior in simulations.
          Different horizons or more complicated prediction models are not studied.
    \item We used the calibrated parameters for SFM and TSFM from~\cite{Jafari2026-TRO} with some experimental tuning.
          The model parameters are not tuned for PSFM and PTSFM in particular. 
          A specialized calibration potentially improves the models' performance
    \item Some participants interpreted the robot's cautious stopping behavior as hesitation and uncertainty from the robot. 
          Thus, the interpretation adds noise to the corresponding reported comfort levels regarding the subjective safety sub-metric.
\end{itemize}

%%%%%%%%%%%%%%%%%%%%%%%%%%%%%%%%%%%%%%%%%%%%%%%%%%%%%%%%%%%%%%%%%%%%%%%%%%%%%%%%%%%%%%%%%%%%%%%%%%%%%%%%%%%%%%%%%%%%%%%%%%%%%%%%%
%%%%%%%%%%%%%%%%%%%%%%%%%%%%%%%%%%%%%%%%%%%%%%%%%%%%%%%%%%%%%%%%%%%%%%%%%%%%%%%%%%%%%%%%%%%%%%%%%%%%%%%%%%%%%%%%%%%%%%%%%%%%%%%%%
%%%%%%%%%%%%%%%%%%%%%%%%%%%%%%%%%%%%%%%%%%%%%%%%%%%%%%%%%%%%%%%%%%%%%%%%%%%%%%%%%%%%%%%%%%%%%%%%%%%%%%%%%%%%%%%%%%%%%%%%%%%%%%%%%
%%%%%%%%%%%%%%%%%%%%%%%%%%%%%%%%%%%%%%%%%%%%%%%%%%%%%%%%%%%%%%%%%%%%%%%%%%%%%%%%%%%%%%%%%%%%%%%%%%%%%%%%%%%%%%%%%%%%%%%%%%%%%%%%%
\section{Conclusion}\label{sec:con:main}
Public spaces host mobile robots and their primary users, the pedestrians.
Therefore, the study of pedestrians' safety during interactions is essential.
In this paper, we review the basic distance-based SFM and the PTTC-based TSFM and introduce their predictive counterparts, PSFM and PTSFM.
Then, we examine their performance in experiments and study the objective and subjective safety of pedestrians when interacting with mobile robots.
The paper statistically analyzes and benchmarks the four SFM variants, considering both objective safety metrics and subjective comfort ratings. 
\par
The results confirm that the PTTC integration improves objective safety and, in fully independent trials, aligns with~\cite{Jafari2026-TRO}.
On the other hand, the prediction integration has a limited impact on most objective safety metrics.
Regarding the subjective safety metrics, qualitative and descriptive trends show that some participants reported improvements in movement smoothness and speed behavior with predictive navigators.
\par
Immediate follow-ups include longer prediction horizons, relaxing the constant velocity assumption, various maximum velocities, and improved parameter calibrations.
Moreover, since the trajectory data is available, our ongoing work focuses on estimating the human-reported comfort levels using a classification/learning technique.
Extending the experiments to mixed crowds with multiple pedestrians and multiple mobile robots roaming the sidewalks is another direction.
In the multi-robot scenario, the agents may act individually, fully cooperate, or partially cooperate.
\par
% \section*{Acknowledgments}
% This should be a simple paragraph before the References to thank those individuals and institutions who have supported your work on this article.
% {\appendix[Proof of the Zonklar Equations]
% Use $\backslash${\tt{appendix}} if you have a single appendix:
% Do not use $\backslash${\tt{section}} anymore after $\backslash${\tt{appendix}}, only $\backslash${\tt{section*}}.
% If you have multiple appendixes use $\backslash${\tt{appendices}} then use $\backslash${\tt{section}} to start each appendix.
% You must declare a $\backslash${\tt{section}} before using any $\backslash${\tt{subsection}} or using $\backslash${\tt{label}} ($\backslash${\tt{appendices}} by itself
%  starts a section numbered zero.)}


\ifCLASSOPTIONcaptionsoff
  \newpage
\fi

	\bibliographystyle{IEEEtran}
	\bibliography{Ref}

\vfill

\end{document}